\documentclass{CVM}

\usepackage{times}
\usepackage{epsfig}
\usepackage{graphicx}
\usepackage{amsmath}
\usepackage{amssymb}
\usepackage{booktabs}  
\usepackage{bm}
\usepackage{stfloats}

\usepackage[pagebackref=true,breaklinks=true,letterpaper=true,colorlinks,bookmarks=false]{hyperref}

\CVMsetup{
type      = {Research Article},
doi       = {s41095-0xx-xxxx-x},
title     = {Neural-Polyptych: Content Controllable Painting Recreation for Diverse Genres},
author    = {Yiming Zhao*$^{1}$, Dewen Guo*$^{1}$, Zhouhui Lian$^{1}$,
Yue Gao$^{2}$,
Jianhong Han$^{3}$,
Jie Feng$^{1}$,
Guoping Wang$^{1}$, Bingfeng Zhou$^{1}$, and Sheng Li$^{1}$ \cor{}\\
},
runauthor = {Yiming Zhao, Dewen Guo, et al.},
abstract  = {
To bridge the gap between artists and non-specialists, we present a unified framework, Neural-Polyptych, to facilitate the creation of expansive, high-resolution paintings by seamlessly incorporating interactive hand-drawn sketches with fragments from original paintings. We have designed a multi-scale GAN-based architecture to decompose the generation process into two parts, each responsible for identifying global and local features. To enhance the fidelity of semantic details generated from users' sketched outlines, we introduce a Correspondence Attention module utilizing our Reference Bank strategy. This ensures the creation of high-quality, intricately detailed elements within the artwork. The final result is achieved by carefully blending these local elements while preserving  coherent global consistency. Consequently, this methodology enables the production of digital paintings at megapixel scale, accommodating diverse artistic expressions and enabling users to recreate content in a controlled manner. We validate our approach to diverse genres of both Eastern and Western paintings. Applications such as large painting extension, texture shuffling, genre switching, mural art restoration, and recomposition can be successfully based on our framework.
},
keywords  = {painting recreation, art, correspondence attention, generative neural network},
copyright = {The Author(s)},
}

\begin{document}


\maketitle

    \begin{figure}[b] \vskip -1mm
    \small\renewcommand\arraystretch{1.3}
        \begin{tabular}{p{80.5mm}} \toprule\\ \end{tabular}
        \vskip -4.5mm \noindent \setlength{\tabcolsep}{1pt}
        \begin{tabular}{p{3.5mm}p{80mm}}
    $1\quad $ & Peking University, Beijing, 100871, China. E-mail: Y. Zhao 1900012759@pku.edu.cn; 
    D. Guo guodewen@pku.edu.cn; Z. Lian lianzhouhui@pku.edu.cn; J. Feng feng\_jie@pku.edu.cn; B. Zhou cczbf@pku.edu.cn; G. Wang wgp@pku.edu.cn; S. Li lisheng@pku.edu.cn. 
    Y. Zhao and D. Guo contributed equally to this work. \\
    $2\quad $ & Microsoft Research Asia, Beijing, 100080, China. E-mail: yuegao@microsoft.com.\\
    $3\quad $ & Huawei Technologies Ltd., Shenzhen, 518028, China. E-mail: hanjh@pku.edu.cn.\\

\end{tabular} \vspace {-3mm}   \end{figure}


\section{Introduction}

\label{sec:intro}

Creating artistic paintings is a profound human skill, serving to convey creative thinking and reflect the societal context of different eras at that moment. Examples range from Chinese masterpieces such as \emph{Along the River During the Qingming Festival} to monumental works like the \emph{Sistine Chapel ceiling frescoes}, which narrate the lives of Moses and Christ. 
Historically, artists have established diverse painting genres, a tradition that persists.
Nowadays, computer-aided art creation offers the opportunity to bridge the gap between artists and non-specialists
\cite{hertzmann2020visual,liu2020towards,xue2021end,fu2021multi}.
Specifically, deep learning-based generative models have made progress in a variety of image translation and genre style transfer tasks in recent years
\cite{azadi2018multi,chen2018gated,xu2021drb,liu2019swapgan,liu2021improved}.
However, their controllability is often limited, constraining the generation of specific stylistic content according to input images. This restricts users from freely designing content to bring their artistic visions to life.

Sketch-based image generation has shown promise in synthesizing photo-realistic images from hand-drawn sketches  \cite{zhu2017unpaired,park2020contrastive,isola2017image,mirza2014conditional,chen2009sketch2photo}.
Nevertheless, manually crafting digital paintings can be labor-intensive, rendering the process unsuitable for non-specialists. Even skilled artists find it time-consuming to create pastiches from scratch. Moreover, the fixed canvas size adds constraints. 

\begin{figure}[t!]
    \centering
    \includegraphics[width=\linewidth]{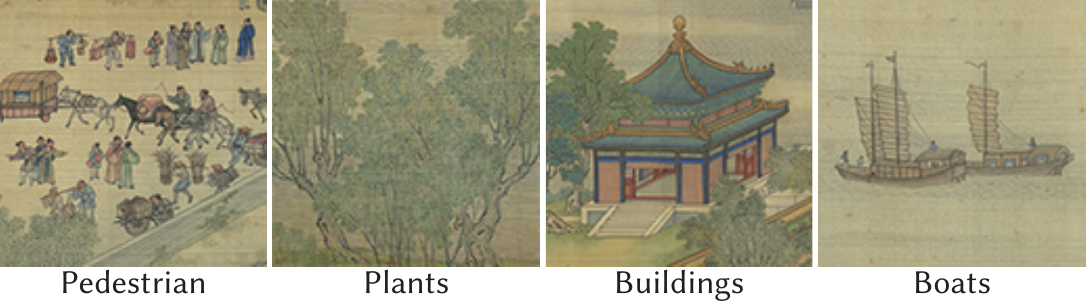}
    \caption{Diverse components with detail from an artwork.}
    \label{fig:categories}
\end{figure}

A feasible solution involves learning-based methods using existing artworks to compose novel images in alignment with original genres. Yet, challenges persist:
\begin{itemize}
\item\emph{Scarcity of painting data:} previous controllable image generation tasks are often guided by specific explicit attributes \cite{karras2020analyzing,karras2019style,karras2017progressive,choi2020stargan,liu2019few}.
Unlike attributes in facial datasets such as CelebA~\cite{liu2015faceattributes}, art painting datasets with specified attributes are scarce.
Robust networks often demand substantial,  datasets with similar content for training, limiting their capacity to handle high-resolution images with minimal annotation. 

\item\emph{Meaningful objects and consistency:} while previous generative frameworks excel in manipulating content composition, size, and aspect ratio, generating meaningful components like pedestrians, buildings, and boats within a coherent narrative context remains challenging (e.g., see Fig.~\ref{fig:categories}). Depicting narrative scenes necessitates adherence to semantic rules and artistic coherence.

\item\emph{Image size constraints:} existing works commonly rely upon public datasets containing lower-resolution images ($256 \times 256$) or, at best, the $1K$ resolution achieved by Progressive GAN~\cite{karras2017progressive} for facial data.
Although solutions like TileGAN \cite{fruhstuck2019tilegan} address non-stationary texture synthesis, unconditional GANs lack convergence stability and control over generated outcomes.
\end{itemize}

\begin{figure*}
    \centering
    \includegraphics[width=\linewidth]{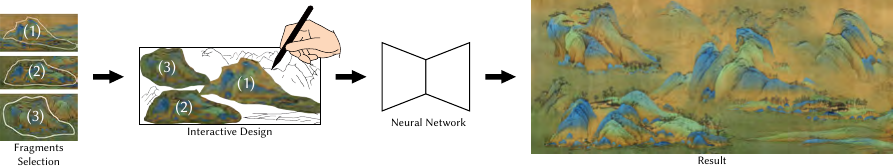}
    \caption{Workflow using our approach. Firstly, we choose specific fragments of interest from the original painting earmarked for recomposition. Subsequently, the user provides interactive sketches, allowing for unconstrained design inputs. Using a neural network, our method transforms these inputs into visually captivating digital paintings.}
    \label{fig:intro-pipeline}
\end{figure*}

In this paper, we develop a learning-based painting recreation approach, \emph{Neural-Polyptych}, which can synthesize large-scale high-definition paintings in a content-controllable manner. Our approach draws inspiration from the artistic technique utilized in churches and cathedrals, referred to as a \emph{polyptych}. A polyptych involves assembling various large panels to compose a grand artwork, particularly suitable for portraying intricate narrative scenes. 

Our approach is dedicated to empowering users to engage in the creative recomposition of original painting patches while seamlessly enjoying the recreation process through interactive sketches. Given the challenge of learning artistic styles from sparse training data available in the original paintings, we present a GAN-based neural network with an encoder-decoder structure to generate paintings in the desired style from the sketch input.
To address the task of generating meaningful objects within paintings, we treat it as a nonlinear optimization problem with boundary conditions~\cite{karras2019style},
employing a novel subspace linear projection technique termed \emph{Reference Bank}. This strategy contributes a gradient prior to the optimization process, aiding in maintaining alignment with the underlying manifold. We further integrate this linear projection into our inference pipeline using a mechanism known as Correspondence Attention (CA), which effectively bridges the gap between the intricate latent space and the linear projection. This technique ensures the separation of distinct reference features to avoid feature mixtures.

Overall, our contributions can be summarized as follows:
\begin{itemize}
    \item We propose a novel content-controllable approach for painting recreation, facilitating the interactive reconstitution of diverse large-scale artworks from pre-existing sources. The versatility of our method extends across various genres, encompassing both Chinese and Western painting traditions. Its workflow is outlined in Fig.~\ref{fig:intro-pipeline}.
    \item We develop an end-to-end generative adversarial net (cGAN) that serves as a unified model, enabling both painting patch splicing and content customization. Leveraging a strategic divide-and-conquer technique for sketch-to-painting translation, our approach generates high-quality, high-definition results.
\end{itemize}

\section{Related Work}
\label{sec:relatedwork}

\subsection{Adversarial Learning}
In the field of image synthesis, adversarial learning can usually achieve better results than the classic auto-encoder~\cite{vincent2010stacked} model.
A generative adversarial network (GAN)~\cite{goodfellow2014generative} usually consists of a generator and a discriminator. The generator aims to learn the distribution of the given data sets, and the discriminator is trained to distinguish synthesized data from real data. The two parts iteratively improve each other via adversarial training. The optimization gradient computed at the discriminator is also passed to the generator for optimization. However, the discriminator is not used in the inference pipeline in the application. The discriminator exists to provide assistance to the generator. 

A standard GAN is optimized only using the loss computed from the discriminator. It is trained in an unsupervised learning fashion without explicit guidance, such as L1 loss for the generator.
Relying only on the GAN loss may cause severe mode-collapse problems~\cite{srivastava2017veegan}. Intuitively, a GAN trained with supervision terms naturally eliminates mode-collapse. DCGAN~\cite{radford2015unsupervised} shows the possibility of combining a supervised CNN and adversarial learning, while iGAN~\cite{zhu2016generative} pioneered interactive image editing using a GAN. For high-quality results, we also adopt a strategy of combining supervised and adversarial learning.
Additionally, several works like \cite{skorokhodov2021aligning,lin2021infinitygan} solve the problem of arbitrary-sized image generation, making it possible to produce extremely high-resolution images using GANs.

\subsection{Image Translation}
Existing image-to-image translation includes several kinds of tasks defined on different datasets. 
At present, there are mainly two types of datasets: facial images and general images in the wild.

Progressive GAN (ProGAN) \cite{karras2017progressive} introduces a fundamental learning strategy of progressively adding layers to both the generator and discriminator. This unconditional GAN is trained on an aligned high-definition ($1,024\times 1,024$) facial image dataset.
ProGAN aims to tackle unconditional image-generation tasks.
Therefore, it is also suitable for less control-demanding tasks such as non-stationary texture generation \cite{fruhstuck2019tilegan}.
StyleGAN \cite{karras2019style} provides a latent code projection solution to borrow styles from target style images. By cooperating with a mapping network composed of several AdaIN-Conv blocks, the synthesis results outperform the progressive generator baseline.
StarGAN v2~\cite{choi2020stargan} proposes a multi-domain image-to-image translation method while satisfying diversity and scalability over multiple domains.
The key consideration of the proposed method is a transformation in latent space, which formulates uniform mapping from randomly selected latent codes to style codes.
Although facial images are semantics-demanding, they are spatially aligned.
However, the complexity of images in the wild or artistic paintings is far greater than for faces.
They usually contain several kinds of meaningful objects,
making such image-to-image translation more challenging.

Pix2PixHD~\cite{isola2017image, wang2018high} proposes a high-resolution image-to-image generation framework using the multi-scale training methodology with the PatchGAN~\cite{isola2017image} discriminator.
We borrow the patchGAN discriminator for its superior performance in image-translation tasks.

SPADE~\cite{park2019semantic} proposes a denormalization method for semantic image synthesis using semantic segmentation masks. The semantic information is better preserved against common normalization layers such as instance normalization~\cite{ulyanov2016instance,huang2017arbitrary}, batch normalization~\cite{ioffe2015batch}, and layer normalization~\cite{ba2016layer,xu2019understanding}. 
As claimed in the literature, those normalizations tend to lose the semantic features learned from the previous layers when applied to uniform or flat input images. Thus, the authors propose a solution using the layout of the semantic input for modulating the activations in normalizations.
Since SPADE is a state-of-the-art image translation framework, we conduct extensive comparative experiments in Sec.~\ref{sec:results}.

CUT~\cite{park2020contrastive} aims to provide unpaired image-to-image translation by contrastive learning. The authors suppose an arbitrary patch in the output should reflect the content of the counterpart in the corresponding input image. To guide the learning process of the network, a positive patch and several negative patches are set for contrastive learning. The learning process is pushed forward by a novel patchwise contrastive loss, which naturally reflects the mutual information between the selected patches.

The methods mentioned above are effective on publicly available image data sets. However, they are difficult to apply to artistic scenarios, and large-scale image generation is generally infeasible.

\subsection{Style Transfer}
Style transfer is a classical problem that has been fully discussed in the field of non-photorealistic rendering over decades. Recent data-driven methods explicitly carry out manipulations in latent space by extracting content and style features from an image~\cite{gatys2016image,li2017demystifying,jing2019neural}.

Neural style transfer usually utilizes a convolution neural network named VGG~\cite{simonyan2014very}, whose parameters are pre-trained on ImageNet~\cite{deng2009imagenet,krizhevsky2017imagenet}, to explicitly extract style representations from specific convolutional layers.
Wang et al.~\cite{wang2020collaborative} propose universal style transfer based on the observation that  GPU memory limits universal style transfer for large-scale images.
Thus, the authors present a new knowledge distillation method to reduce the convolutional filters in the encoder-decoder-based deep architectures.
Kim et al.~\cite{kim2020deformable} redefine style transfer in two ways: geometry and texture. The geometric information encoded in photorealistic images is also fundamental to visual style. The image deformation is firstly specified by a set of source key points generated using several off-the-shelf frameworks, followed by automatic warping according to the key points.

In addition to style, there is something else worth extracting from the rich collection of relics and artworks: the original authors‘ creative genres, which we discuss later in detail.
Existing style transfer methods cannot allow users to modify any semantic features or content in the target images, i.e.\ interactive design is not allowed. Therefore, style transfer is unsuitable for our task.

\section{Method}
\label{sec:method}

\begin{figure*}
    \centering
    \includegraphics[width=\linewidth]{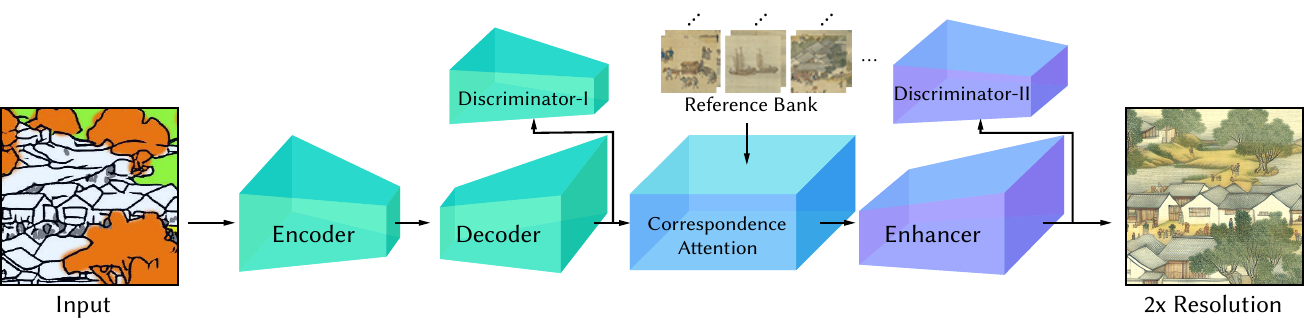}
    \caption{Overview of our network architecture.}
    \label{fig:net}
\end{figure*}

\subsection{Architecture}
Our neural network for painting generation has two main parts,  a fully convolutional network (FCN) with a classic encoder-decoder structure as the master branch and a super-resolution enhancer with a proposed correspondence attention (CA) module.
In the first place, the encoder $E$ takes in the sketch images as inputs and produces preliminary results under the supervision of Discriminator-I $D$, which takes the outputs from the decoder for loss computation to ensure the quality of the generated results.
After that, embedding vectors are extracted using pre-trained VGG19 from both the produced outputs and the reference bank (RB), which are used to compute spatial and channel attention in the CA module to build hybrid features. 
Finally, under the supervision of discriminator-II, the enhancer $H$ takes in the outputs of CA and generates high-resolution results with high-quality details. An overview of our network architecture is illustrated in Fig.~\ref{fig:net}. Details of each module are described in the following subsections.

\begin{figure*}[t]
    \centering
    \includegraphics[trim={0 0 2cm 0},clip, width=\linewidth]{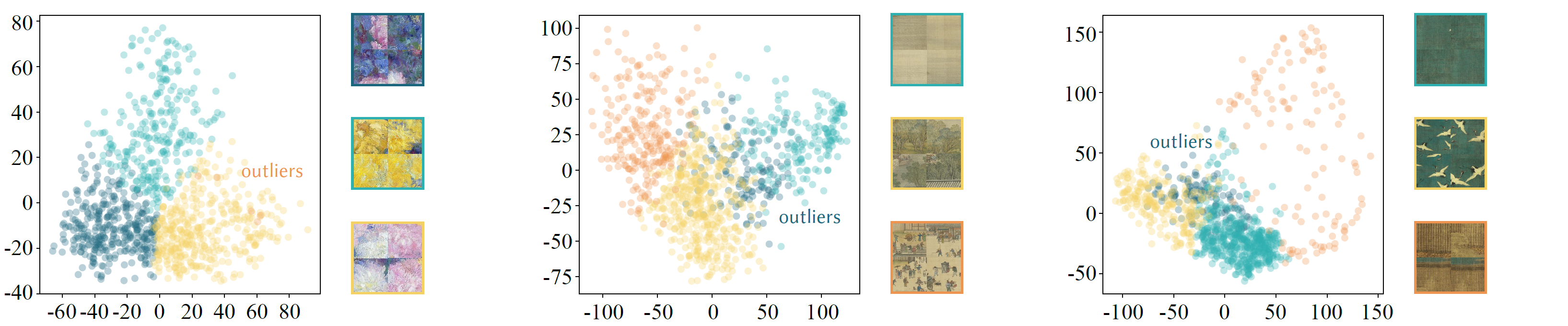}
    \caption{Visualization of clustered results of the reference bank. Three representative examples of the category are showcased on the right side of each dot chart. Dots' colors in the chart match the frame color of the corresponding representative example. Categories with few samples $\left(< N\right)$ are treated as outliers.}
    \label{fig:cluster}
\end{figure*}

\begin{figure}[t]
    \centering
    \includegraphics[width=\linewidth]{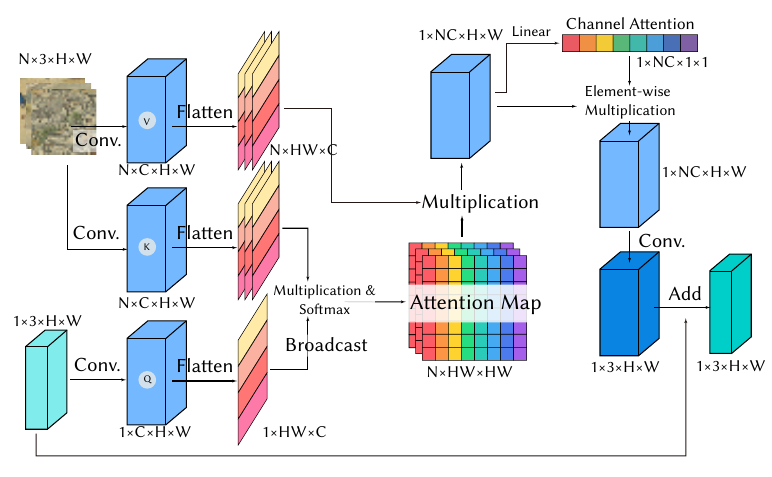}
    \caption{Correspondence Attention.}
    \label{fig:CA}
\end{figure}

\subsection{Encoder-Decoder Structure}
Variational autoencoders (VAEs) and generative adversarial networks (GANs) commonly employ an encoder-decoder structure. As depicted in Fig. \ref{fig:net}, the encoder maps data to a latent space, effectively capturing and encoding essential features from the input data. The decoder, on the other hand, takes this condensed representation and converts it into the desired output format, ultimately generating a reconstructed image.
To train the model, the discriminator plays a crucial role by minimizing the error between the reconstructed image and the original image within the pixel space. This adversarial training process helps enhance the fidelity of the generated images.

\subsection{Reference Bank}
\label{sec:refbank}
To recover content details of some sophisticated artworks, it is not enough to simply use pixel-to-pixel architecture when generating final results. So, we designed the reference bank (RB) to ensure that the outputs of our network contain sufficient high-quality artistic details. 

Existing conditional GANs usually introduce simple functions such as $L1$ loss as quality metrics.
Such loss functions easily converge due to their natural linearity.
However, it is quite challenging when the data provided does not have a uniform composition.
Our RB module can be regarded as the projection from an entangled manifold of optimization to a set of linear subspace approximations.
We formulate it by mapping to specific categories of references.

To build the reference bank, firstly, we decompose the original painting into multi-resolution pieces, e.g., 64$\times$64, 128$\times$128, and 256$\times$256.
Next, we feed the samples into a pre-trained VGG19 network to obtain the corresponding embeddings from certain (\texttt{conv3\_3} and \texttt{conv4\_2}, to be exact) hidden layers.
After that, the feature maps are concatenated to a $4^{th}$ order tensor for hierarchical clustering \cite{ziegel2003elements}. In this way, we divide the reference patches into $n$ different categories, which can then guide parameter learning in the following CA module.  PCA-reduction of the clustered results is demonstrated in Fig.~\ref{fig:cluster}.

Given an input at a certain scale, all the painting pieces in different categories are adjusted to the same size as the input using zero-padding and equal-scale resizing. In this way, we ensure that objects in the components will not be stretched unreasonably.

\subsection{Correspondence Attention}
\label{sec:CA}
Artwork consists of not only stylized texture but also a variety of meaningful components, as shown in Fig.~\ref{fig:categories}. When generating paintings from user sketches, simply recovering textures of the same style and shape will only produce blurry, low-quality results. Therefore, we should consider small components and try to produce high-quality output by arranging and rendering them properly under the guidance of input sketches. Different input sketches usually focus on different expected components, so it is natural to use a visual attention mechanism in the model to adapt to various input sketches.

The visual attention mechanism bridges the gap between long-range dependencies, relaxing the narrow perceptive fields of CNNs.
Therefore, we adopt similar thoughts to construct the cross-domain dependencies between intermediate features and reference patches randomly selected from RB. As the patches from RB have different kinds of main content in different positions, we naturally consider both spatial and channel attention in our correspondence attention (CA) module, as illustrated in Fig.~\ref{fig:CA}.

For spatial attention, CA takes the output ($\bm{X}$) from the decoder as input for computing the attention map using the reference patches ($\bm{\text{Ref}_i}, i = 1, \dots N$, resized as the shape of $\bm{X}$) from RB:
\begin{align}
    Q & = \text{Flatten}(\text{Conv}_q(\bm{X})),\\
    K_i & = \text{Flatten}(\text{Conv}_k(\bm{\text{Ref}_i})),\\
    V_i & = \text{Flatten}(\text{Conv}_v(\bm{\text{Ref}_i})),\\
    \text{AttnMap}_i & = \text{Softmax}(Q^T \cdot K_i),\\
    \text{Mid}_i & = (\text{AttnMap}_i \cdot V_i^T)^T,
\end{align}
where $\text{Conv}_*$ refers to convolution layers, and $Q,K_i,V_i \in \mathbb{R}^{C \times HW}$ refer to query, key, and value features, respectively. We concatenate all  $\text{Mid}_i \in \mathbb{R}^{C \times HW}$ in the channel dimension to generate the intermediate feature $\text{Mid} \in \mathbb{R}^{NC \times H \times W}$.
Note that the inputs of a network may contain different kinds of main content, so we should pay different attention to the reference patches in RB, which means paying attention to the dimension of reference channels.

Intuitively, since the input of our network can have any size, we can get the channel-wise hidden weights as in SEnet~\cite{8701503}:
\begin{align}
    z &= \text{Avgpool}(\text{Mid}),\\
    W &= \sigma\left(W_2\delta(W_1z)\right),
\end{align}
where Avgpool calculates the mean value for each hidden channel to get $z \in \mathbb{R}^{NC \times 1 \times 1}$,  $\delta$ refers to GeLU activation and $\sigma$ refers to a sigmoid function.

To ensure  stable convergence of the linear layers $W_1, W_2$, we arrange the reference patches in a specific order: we repeatedly sample the reference patch from each category, as mentioned in Sec.~\ref{sec:refbank}, and arrange them as:
\begin{equation}
\begin{split}
    \bold{Ref} = &[\left(\text{Ref}_1 \dots \text{Ref}_k\right), \left(\text{Ref}_{k+1} \dots \text{Ref}_{2k}\right), \dots, \\
    &\left(\text{Ref}_{(n-1)k+1} \dots \text{Ref}_{nk}\right)],
\end{split}
\end{equation}
where $k$ is the number of categories and thus $nk = N$.

Finally, the output of CA can be computed as:
\begin{align}
    \text{Attn} &= \text{Conv}_{1\times 1}\left(W * \text{Mid}\right) \ ,\\
    \text{Out} &= \bm{X} + \lambda * \text{Attn} \ ,
\end{align}
where $*$ indicates element-wise multiplication using a broadcast mechanism, 
Attn has the same shape as $\bm{X}$ and $\lambda$ is a learnable variable.

To this end, we fix the mixing of different samples to make the manifold as linear as possible.

\subsection{Multi-scale GAN Training}
Our multi-scale GAN offers enhanced stability during the training process. Training a GAN can be challenging, often requiring a delicate balance between the generator and discriminator networks. Our multi-scale GAN incorporates a progressive growth element, gradually increasing the complexity of both the generator and discriminator networks. This approach can mitigate problems like mode collapse and training instability, resulting in more reliable training.
In addition, by incrementally introducing fine details to the generated images in two consecutive stages within the multi-scale GAN, our approach becomes highly proficient at preserving intricate local details of the paintings.
A GAN is trained individually for each painting. More precisely, the training process involves utilizing randomly selected image patches from arbitrary locations within the painting as the training samples.
Throughout the training iterations, we closely monitor the losses of both the generator and discriminator. The GAN is considered to have converged when the generator and discriminator losses are stable, as well as when the values of metrics such as FID and IS cease to exhibit significant changes. In other words, the GAN is deemed to have converged when there is a consistent and minimal fluctuation in the aforementioned indicators.

\begin{figure}[t]
    \centering
    \includegraphics[width=\linewidth]{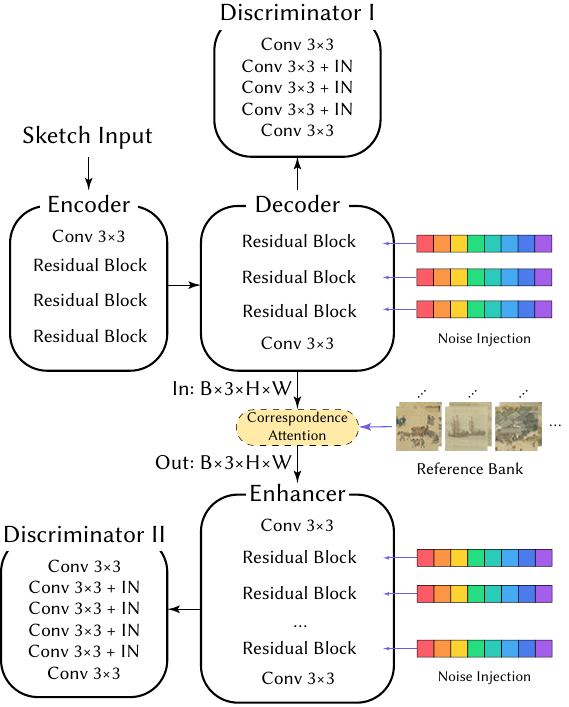}
    \caption{Network architecture including implementation details.}
    \label{fig:network-detail}
\end{figure}

\subsection{Objective Functions}
Given an input sketch image $s\in\mathcal{S}$, its corresponding painting $p\in\mathcal{P}$, and reference patches $\text{Ref}_i, i = 1, \dots N$, we train our approach using the following objectives.

\subsubsection{multi-scale GAN loss}
As shown in Fig.~\ref{fig:net}, multi-scale discriminators take the output images of the decoder and enhancer as input images, respectively; we formulate the multi-scale GAN loss as:
\begin{equation}
\begin{split}
\mathcal{L}_\text{GAN}^I &= \mathbb{E}_{p}\left[\log D_p^I\left(\mathcal{H}\left(p\right)\right)\right]\\
& +\mathbb{E}_{s,p}\left[1-\log D_p^I\left(G^I\left(s\right)\right)\right],
\end{split}
\label{eq:loss_gan_I}
\end{equation}
\begin{equation}
\begin{split}
\mathcal{L}_\text{GAN}^{II} &=\mathbb{E}_{p}\left[\log D_p^{II}\left(p\right)\right]\\
& +\mathbb{E}_{s,p}\left[1-\log D_p^{II}\left(G^{II}\left(s; \text{Ref}_1,\text{Ref}_2\dots\right)\right)\right],
\end{split}
\end{equation}
where $D_p^I$ and $D_p^{II}$ denote discriminator-I and discriminator-II, respectively, while $G^I$ and $G^{II}$ represent the encoder-decoder and enhancer with CA, respectively. $\mathcal{H}\left(p\right)$ in Eq.~\ref{eq:loss_gan_I} is the halving operator which resizes $p$ from $H\times W$ to ${(H}/{2}) \times ({W}/{2)}$.

\subsubsection{Contextual similarity loss} To ensure  content consistency between the network output and real image, we also formulate  contextual similarity loss as:
\begin{equation}
    \mathcal{L}_\text{cx}^{I} = \mathbb{E}_{s,p}\left[\text{CX}\left(V_p\left(G^{I}\left(s\right)\right),V_p\left(\mathcal{H}(p)\right)\right)\right],
\end{equation}
\begin{equation}
    \mathcal{L}_\text{cx}^{II} = \mathbb{E}_{s,p}\left[\text{CX}\left(V_p\left(G^{II}\left(s; \text{Ref}_1,\text{Ref}_2\dots\right)\right),V_p(p)\right)\right],
\end{equation}
where $V_p$ indicates the perceptual feature produced by pre-trained VGG19 and CX is the  contextual loss calculated as  in \cite{mechrez2018contextual}.

\subsubsection{L1 loss} The learning methodology is basically supervised training.
To ensure generation quality, an additional L1 loss is applied to the objective:
\begin{equation}
    \mathcal{L}_{1}^{I}=\mathbb{E}_{s,p}\left[||G^{I}\left(s\right) - \mathcal{H}(p)||\right],
\end{equation}
\begin{equation}
    \mathcal{L}_{1}^{II}=\mathbb{E}_{s,p}\left[||G^{II}\left(s; \text{Ref}_1,\text{Ref}_2\dots\right) - p||\right].
\end{equation}

\subsubsection{Full objective} When training a conditional GAN, L1 loss is the dominant loss to stabilize optimization convergence; thus, the weight for L1 loss is relatively greater than for the others.
The full objective of our training is
\begin{equation}
    \mathcal{L} = \mu_1\mathcal{L}_\text{GAN} + \mu_2\mathcal{L}_\text{cx} + \mu_3\mathcal{L}_{1},
\end{equation}
where $\mu_1, \mu_2, \mu_3$ are hyperparameters defined for the training tasks.

\section{Experiments and Results}
\label{sec:experiments}

\subsection{Setting}

\subsubsection{Implementation Detail}
Our entire framework was trained end-to-end using 4 nVidia A100 graphics cards with 80GB memory for acceleration. We trained the network for 1000 epochs on an individual case, taking approximately 60 GPU hours. However,  inferencing for testing can be carried out on the CPU on a PC, Mac, or any other device supporting PyTorch 1.4.0 or higher. We used the Adam optimizer with learning rates $\lambda_G=0.0005$ for both generators and $\lambda_D=0.002$ for both discriminators. Additionally, we set $\mu_1=0.1$, $\mu_2=1.0$, $\mu_3=10.0$ in the full objective.
The decision on when to conclude the training process is informed by the observed behavior of metrics, including losses, FID, and IS. This approach is chosen over relying on a predefined dataset due to the diverse dimensions of the painting data.
More details of the model architecture are shown in Fig.~\ref{fig:network-detail}.

For 512$\times$512 outputs, the time taken for inferencing is about 20 ms in the first stage and 120 ms (for super-resolution) in the second stage on a single A100 GPU. Our system consumes only 8.6GB of VRAM when generating $512 \times 512$ images (and 34.5 GB for $1024 \times 1024$), which is possible for many consumer GPUs. Note that our model only contains 84.7M trainable parameters, many fewer than for diffusion models.

\subsubsection{Training Data and Training Strategy}
As mentioned above, popular existing data sets are unsuitable for large-scale painting generation due to the unique task defined in this paper.
Therefore, we constructed a data set from publicly available sources, including artworks from both Western and Eastern paintings.
The data we used are  as follows:
\begin{itemize}
    \item \emph{Along the River During the Qingming Festival} is one of the most important artistic creations in ancient China. The painting contains several kinds of objects, such as buildings, plants, pedestrians, rivers, land, hills, etc.
    \item \emph{Rivers and Mountains} takes stone green and other minerals as the main pigment to exaggerate the color and has a certain decorative nature. It is also called `green landscape.'
    In this case, we introduce a plug-and-play simple perspective mapping that can be applied to the input sketches for processing.
    \item \emph{Almond Blossom}, a painting by Vincent van Gogh, the famous post-impressionist painter.
    \item \emph{Chrysanthemums},  by Claude Monet. We included this case in our dataset to demonstrate our shuffling operation, to recreate artistic paintings without any additional sketch. 
    \item \emph{Other modern paintings for further evaluation}. For each painting, we split the content into two parts for training and testing, respectively. 
\end{itemize}

For each artwork in our dataset, we trained a distinct model individually. These models have the capability to generate new samples in the same genre as the original artwork, guided by novel sketch inputs, without the need for generalization across diverse genres. To assemble the training data, we applied data augmentation techniques to the specific artwork,  via random cropping, flipping, and color transformations. Our model was trained using these augmented images for the specific artwork in a collective manner. We supervised the outputs of our decoder and enhancer at resolutions of $256^2$ and $512^2$, respectively.
The training dataset comprised up to 1 million sample patches at resolutions of either $256^2$ or $512^2$ for each painting. In each epoch, we randomly selected 1,000 patches for training; the maximum number of epochs was set to 1,000.

Initially, our dataset consisted of 4 unique images for each painting category. Recognizing the risk of overfitting due to a restricted number of original samples, we employed a comprehensive data augmentation strategy to expand the dataset effectively. This strategy increased the volume to 1 million images per category, post-augmentation.
Data augmentation was conducted using a combination of techniques to introduce a rich variety of alterations that mimic real-world variations. These techniques included geometric transformations such as rotations, flips, and scaling; photometric changes like brightness and contrast adjustments; and synthetic alterations such as noise injection and filter applications. Each technique was carefully calibrated to ensure that the augmented images remained realistic and relevant to the application context.
To further ensure the generalizability of our model, we employed rigorous validation techniques. Specifically, we used a separate validation set that was not subjected to augmentation to closely monitor for signs of overfitting. Additionally, we applied regularization techniques and cross-validation during training; these are well-established methods for preventing overfitting.
Overall, despite the substantial size of our post-augmentation dataset, we have taken meticulous steps to ensure robust training of our model and prevent artificially inflated performance due to augmented data.

\begin{figure*}[p!]
    \centering
    \includegraphics[height=0.95\textheight]{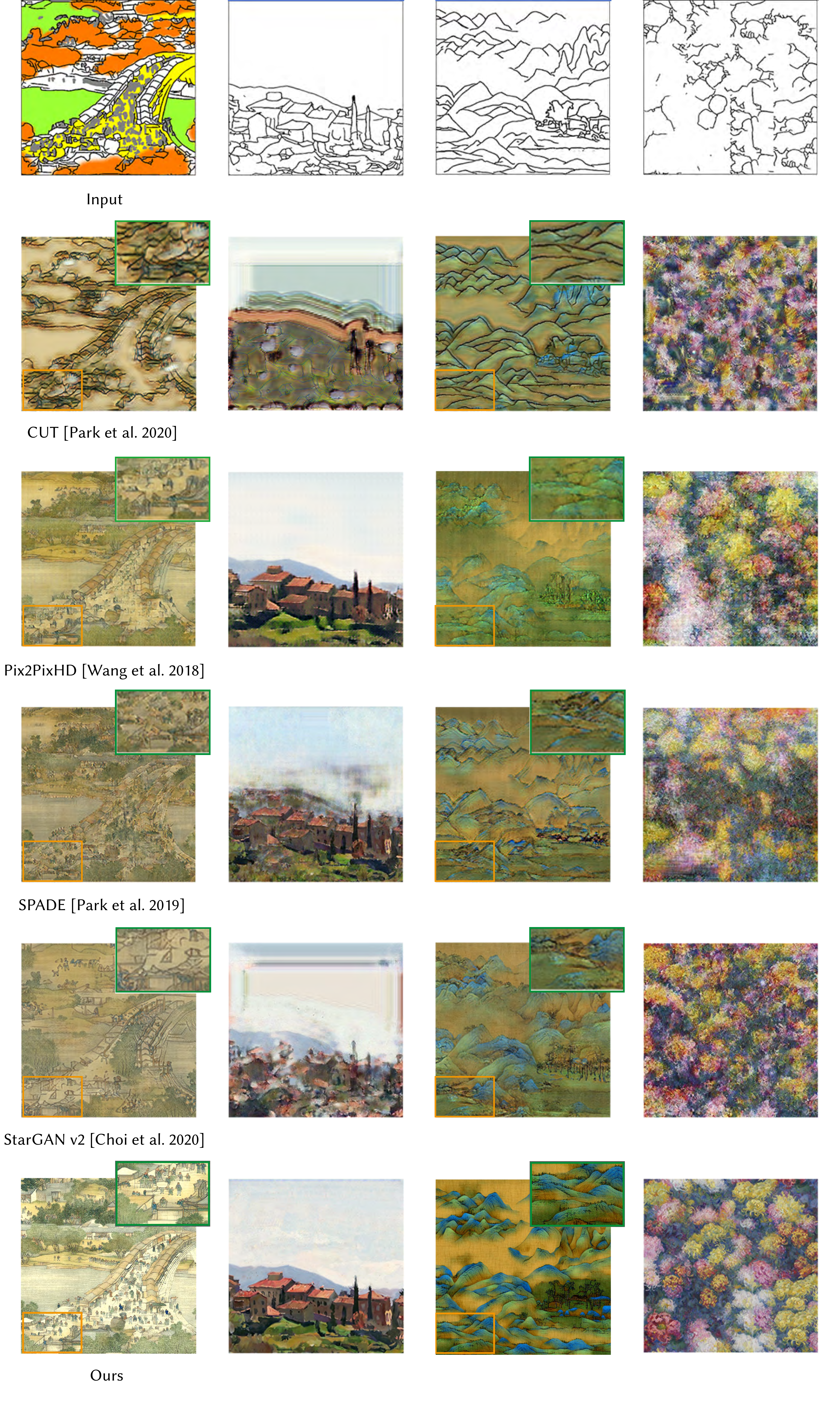}
    \caption{Qualitative comparison of our approach to several state-of-the-art methods to synthesize an image from a sketch. Our approach always shows superior performance over other methods.}
    \label{fig:compare}
\end{figure*}

\subsubsection{User Sketch Generation}
\label{sec:SG}
As shown in Fig.~\ref{fig:intro-pipeline}, users may pick random fragments cropped from the target painting and arrange them arbitrarily on a single canvas. Then, a few sketch lines can be added anywhere on the canvas to guide generation based on the user's desires. All of the above processes can be done in less than a minute. 

With user input, we do sketch extraction on the picked fragments and form the complete sketch together with the added lines.
We chose an off-the-shelf edge detection framework~\cite{xie2015holistically} for sketch extraction and used a learning-based sketch simplification method~\cite{simo2016learning} for noise reduction.

For sophisticated artworks like \emph{Along the River During the Qingming Festival}, simple sketch lines are not enough to generate reasonable results; thus, semantic hints are required in such cases. In practice, users used the Magic Wand Tool in Photoshop to paint color masks on the canvas for semantic region annotation. Since accurate semantic segmentation is unnecessary for our method, such a job can be done quickly and simply.
During the training stages, we used pure sketch lines as input for simple tasks, and sketch lines with color masks for sophisticated tasks.

\subsection{Comparative Results}
\label{sec:results}

\begin{figure*}[tp!]
    \centering
    \includegraphics[width=\linewidth]{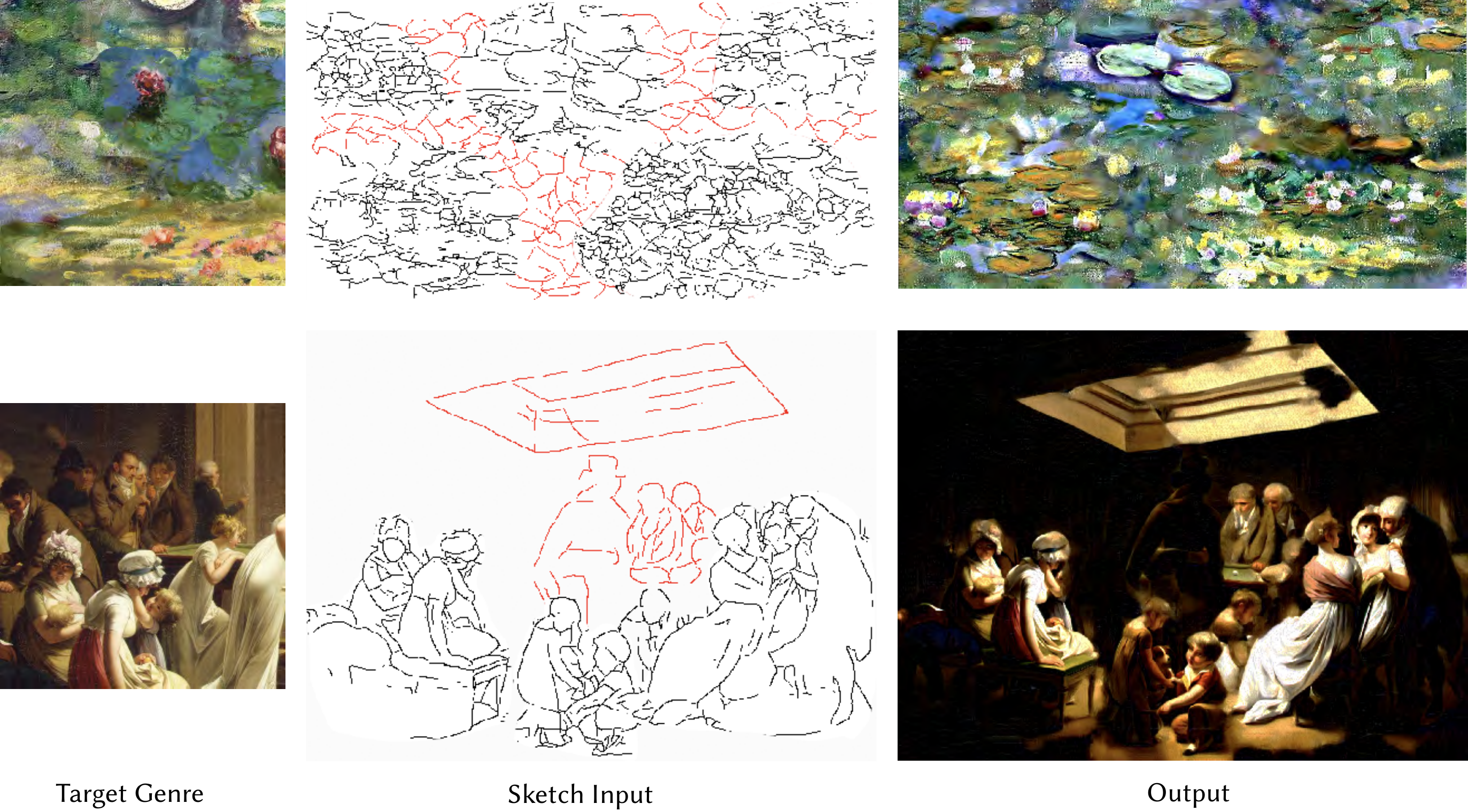}
    \caption{Synthesis of diverse Western art genres. Above: impressionist paintings. Below: group portraits. (Red sketches are hand-drawn input from inexperts.)}
    \label{fig:add-demo}
\end{figure*}

\begin{table}[t!]
\centering
\caption{Quantitative comparison. Our approach provides the best performance in terms of all metrics.}
\resizebox{\columnwidth}{!}{
\begin{tabular}{lccccc}
\toprule
Methods & IS $\uparrow$ & FID $\downarrow$ & LPIPS $\downarrow$  & Pix-Acc $\uparrow$ \\
\midrule
CUT & 1.9755 & 299.9253 & 0.6222  & 0.3418 \\
Pix2PixHD & 1.6235 & 71.3969 & 0.3154  & 0.2973 \\
SPADE & 1.6295 & 116.0550 & 0.3726  & 0.3913 \\
StarGAN v2 & 1.5046 & 189.6963 & 0.5149  & 0.3372 \\
Ours & \textbf{2.0560} & \textbf{41.7846} & \textbf{0.1970} & \textbf{0.8355} \\
\bottomrule
\end{tabular}}
\label{tbl:comparative}
\end{table}

Since most state-of-the-art high-resolution image generation methods like StyleGAN-XL~\cite{Sauer2021ARXIV} are not fit for sketch-level control, and style transfer methods like CAST~\cite{zhang2020cast} have trouble in handling sketch-like input, such methods are excluded from consideration.
Consequently, our experimental strategy embraces content-controllable image translation methods, exemplified by Pix2PixHD~\cite{isola2017image, wang2018high} and SPADE~\cite{park2019semantic}. We furthermore conducted a comparative analysis involving other cross-domain generation methodologies, including CUT and StarGAN v2~\cite{choi2020stargan}. 

We adopted several commonly used evaluation metrics for generative models to evaluate the proposed method:
(i) IS: Inception score~\cite{salimans2016improved} demonstrates the effectiveness of the target models by evaluating the quality and the diversity of the output images from them. (ii) FID: Fréchet Inception Distance~\cite{heusel2017gans} is the distance between the real image and the generated image at the feature level. (iii) LPIPS: Learned Perceptual Image Patch Similarity~\cite{zhang2018unreasonable} evaluates the distance between two image patches. (iv) Pix-Acc: evaluates pixel-level accuracy to reflect the performance of a model directly.

Our approach outperforms the above methods in all evaluation metrics, as listed in Table~\ref{tbl:comparative}.
In addition to the quantitative evaluations, Fig.~\ref{fig:compare} shows results from the test dataset. Our approach can produce more realistic results.

To show the superiority of our approach, we conducted more experiments on Western genres beyond the examples already included above. We took impressionist paintings and group portraits as examples, as shown in Fig.~\ref{fig:add-demo}. Our approach can synthesize desired paintings close to the target.

\subsection{Ablation Study}
\label{ablation study}

We conducted an ablation study on the CA module and multi-scale structure on \emph{Along the River During the Qingming Festival} dataset.  Quantitative results in Table~\ref{tbl:ablation} demonstrate that each part contributes to the final performance.

\begin{figure*}[tp!]
\includegraphics[width=\linewidth]{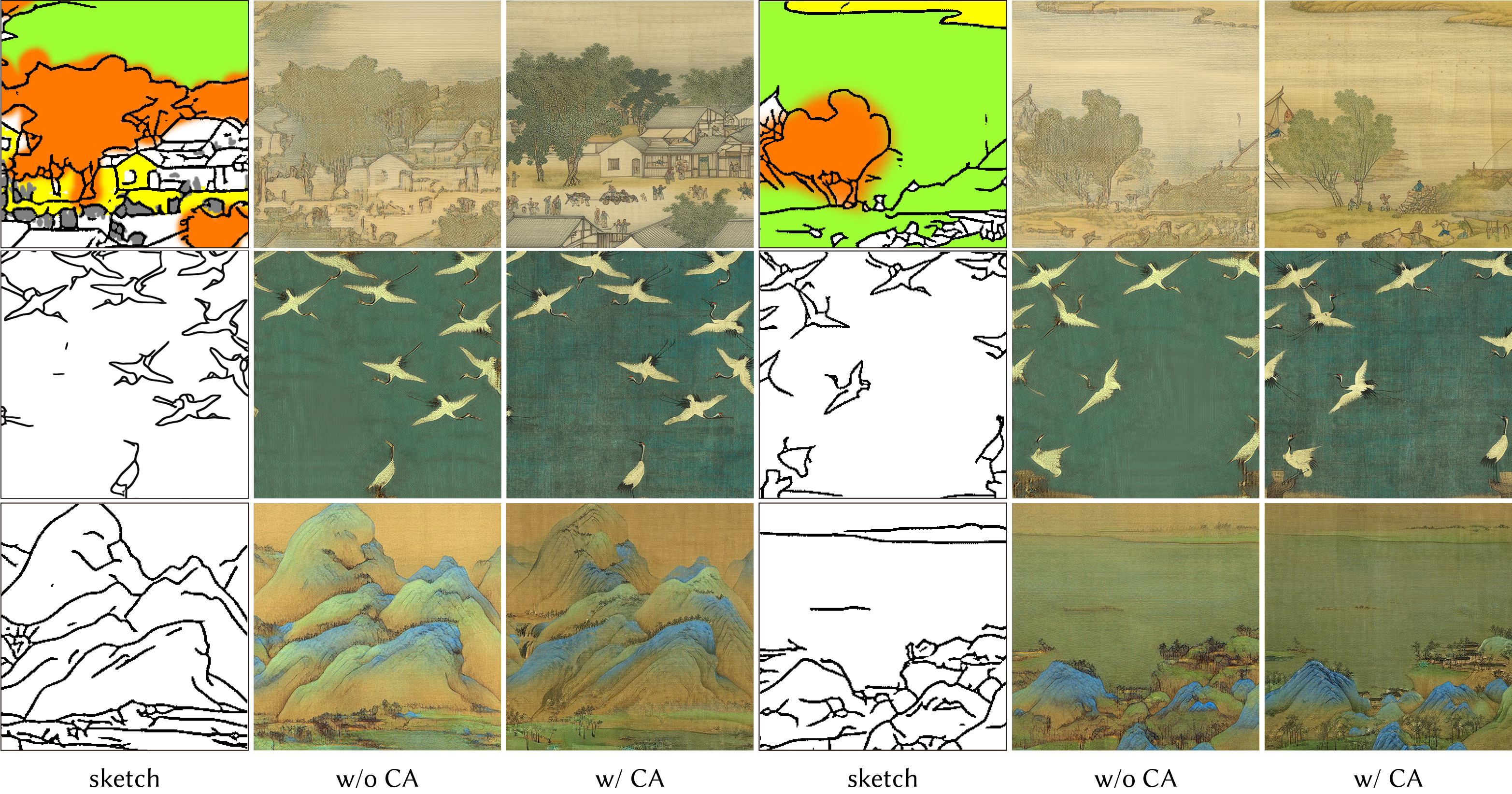}
\caption{Ablation study: with and without Correspondence Attention. High-quality detail is enhanced by the CA module.}
\label{fig:ablation-attention}
\end{figure*}

\begin{figure*}
\includegraphics[width=\linewidth]{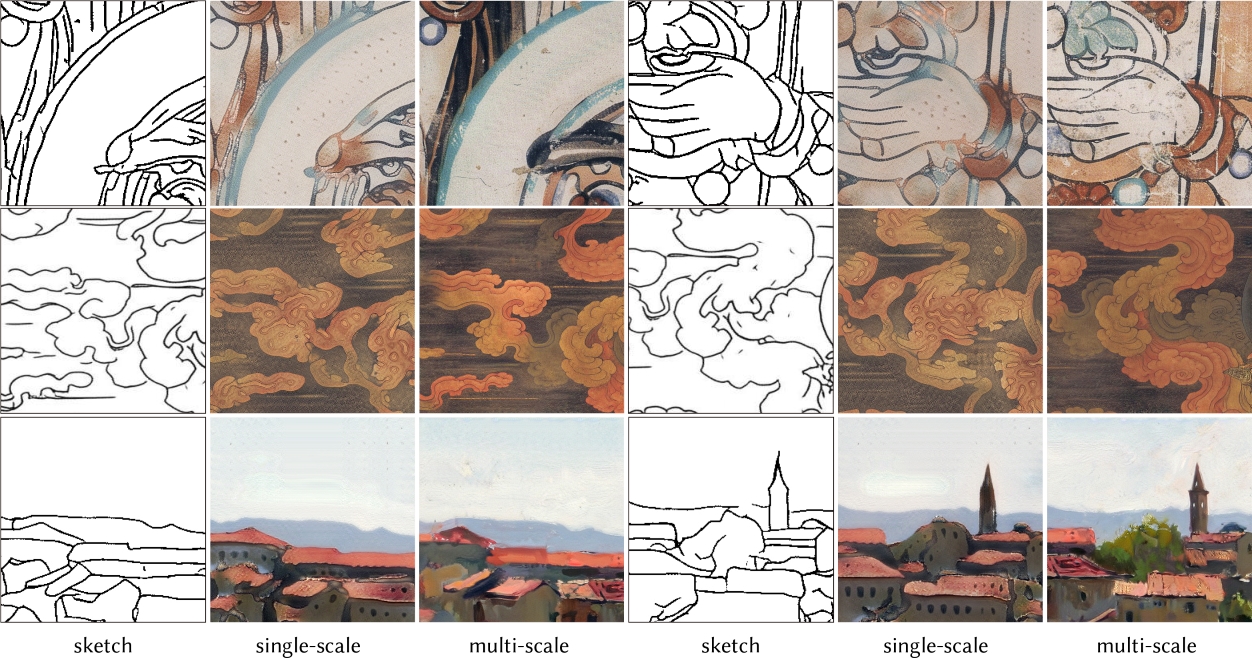}
\caption{Ablation study: single-scale versus multi-scale. Using multi-scale can enhance detail to better match the original genre.}
\label{fig:ablation-scale}
\end{figure*}

In addition, we also conducted ablation studies on several types of art genres to validate the effectiveness of correspondence attention (CA) and the multi-scale strategy.
Figs.~\ref{fig:ablation-attention},~\ref{fig:ablation-scale} demonstrate that both CA and multi-scale strategy can significantly improve the quality of details of the synthesized paintings. Although the models trained without CA can also produce the desired overall appearance, CA provides references for the second-scale generations, giving higher-level hints of visual features noticeable in the boxed regions.

\begin{table}[t]
\centering
\caption{Quantitative evaluation of ablation studies. CA = Correspondence Attention, MS = multi-scale. w/o = without.}
\resizebox{\columnwidth}{!}{
\begin{tabular}{lccccc}
\toprule
Model & IS $\uparrow$ & FID $\downarrow$ & LPIPS $\downarrow$  & Pix-Acc $\uparrow$ \\
\midrule
w/o CA & 1.9450 & 125.8350 & 0.3100 & 0.7575 \\
w/o MS & 1.9296 & 76.6913 & 0.3460 & 0.8275 \\
CA+MS & \textbf{2.0560} & \textbf{41.7846} & \textbf{0.1970} & \textbf{0.8355} \\
\bottomrule
\end{tabular}}
\label{tbl:ablation}
\end{table}

\section{Inexpert User Study}
\label{user}

\begin{figure*}[tp]
    \centering
    \includegraphics[height=0.95\textheight]{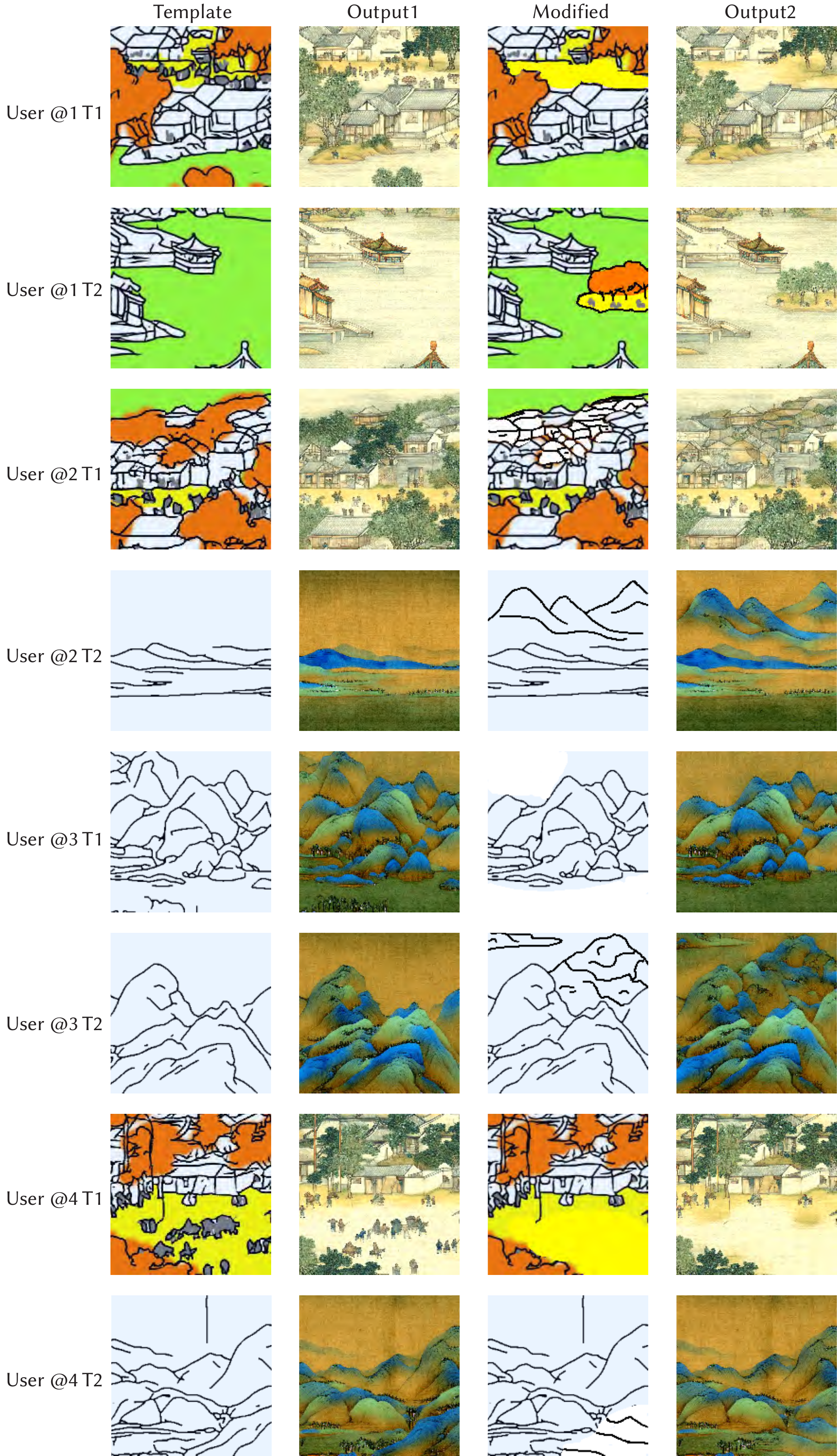}
    \caption{Outcomes from interactive inputs from inexpert users. The two columns on the left showcase the portion of initially supplied templates alongside their corresponding generated outcomes. In contrast, the two rightmost columns depict instances where users engaged in active reprocessing of their initial sketch inputs, resulting in the final outputs.}
    \label{fig:user-result}
\end{figure*}

The sketch inputs utilized in our model are automatically extracted from artistic paintings through edge detection and sketch simplification techniques during the training phase. This process is entirely automated. Through data augmentation and noise injection, the model's generalization capabilities are enhanced during training, enabling it to effectively manage interactive sketch inputs from inexpert users during the inferencing phase. For intricate artworks teeming with details, manual specification of semantic masks becomes necessary to guide the model in generating content for specific regions. This task can be accomplished expeditiously (in approximately 5--10 minutes) using readily available drawing tools like Photoshop or other online platforms.

To evaluate the effectiveness of our approach in processing inputs from the general user, we developed a demonstration system. This system accepts user inputs, either through uploading or real-time drawing, and subsequently generates final paintings using our method. Recognizing that users may lack professional drawing skills, the system offers templates for each artistic genre. These templates are excerpts from original masterpieces, allowing users to interactively modify the sketches. Ultimately, the system leverages these modified inputs to generate paintings guided by the provided sketches.

We extended invitations to several inexpert users to assess our method through the above system. A selection of results is shown in Figure~\ref{fig:user-result}. These outcomes encompass diverse input types, where users have erased, added, or altered sections of the template sketches. In all instances, the system adeptly responds, producing unique outputs reflective of the modifying inputs. The resulting paintings boast remarkable fidelity and contextual coherence, validating the capability of our proposed approach to effectively process inputs from general users. In practical usage, the styles of user sketches vary, which, in turn, impacts the generated content. More precisely, complex sketches with abundant details offer better guidance to our model, helping to achieve high-fidelity domain transfer.  Contrariwise, overly simple and undetailed sketches can confuse the model, making it uncertain about what to generate in specific areas, resulting in less satisfactory outcomes.
As part of our ongoing efforts, we will make our system available online, encouraging broader user participation and soliciting feedback to iteratively enhance usability.

\section{Diverse Applications}
\label{sec:application}
Besides painting recreation, our framework can handle a variety of applications. We show large painting extension, texture shuffle, and genre switch to demonstrate our framework's capability.

\begin{figure*}[t]
    \centering
    \includegraphics[width=\linewidth]{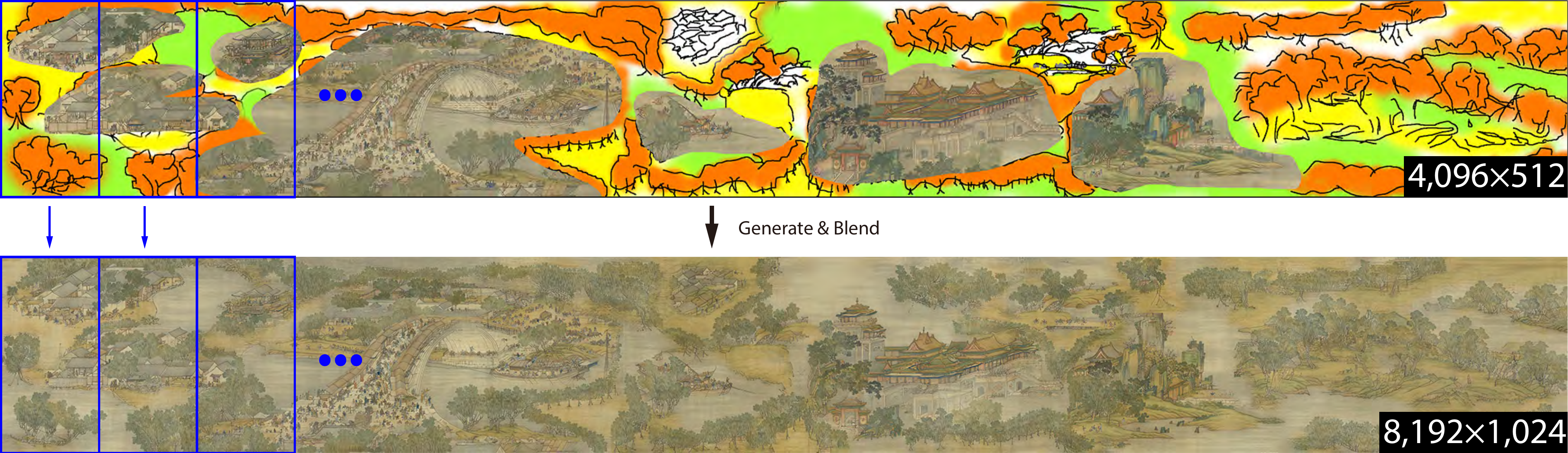}
    \caption{Large-scale painting generation. Given a large-scale canvas, our approach decomposes the canvas into overlapping patches for linear blending. Once generated, all patches are blended into a large-scale overall painting with a resolution twice the width and height of the input canvas.}
    \label{fig:qingming}
\end{figure*}

\subsection{Large-scale Painting Generation}
To generate large-scale paintings using a limited memory configuration,
our approach allows users to decompose the sketches into pieces to generate them individually before blending the local results into a complete painting. In the case demonstrated in Fig.~\ref{fig:qingming}, we provide a canvas at $4K\times 0.5K$ resolution, where the user can recompose and edit with interactive sketches.
After recreation, our approach decomposes the canvas into pieces where two adjacent patches share the overlapping regions of $0.25K\times 0.5K$ resolution. The overlapping regions are set for linear blending.
After generation for all patches, we blend the generated results into a full painting with minimal boundary artifacts.

The pixel values in the overlapping region are weighted sums of the counterparts on both sides.
For example,  for left-right blending,
with resolution of the overlapping region is $W\times H$,
for a pixel $P\left(i, j\right)$ in the overlapping region, its value is decided by
\begin{equation}
    P\left(i,j\right) = \frac{i}{W}P_l^{i,j} + \frac{W-i}{W}P_r^{i,j},
\end{equation}
where $P_l^{i,j}$ and $P_r^{i,j}$ are its counterparts  in left and right patches respectively. When generating an arbitrary large-scale painting,  GPU memory consumption is about 8.6GB for a $512^2$ patch size and about 34.5GB for a $1024^2$ patch size.

\begin{figure}[t]
    \centering
    \includegraphics[width=\linewidth]{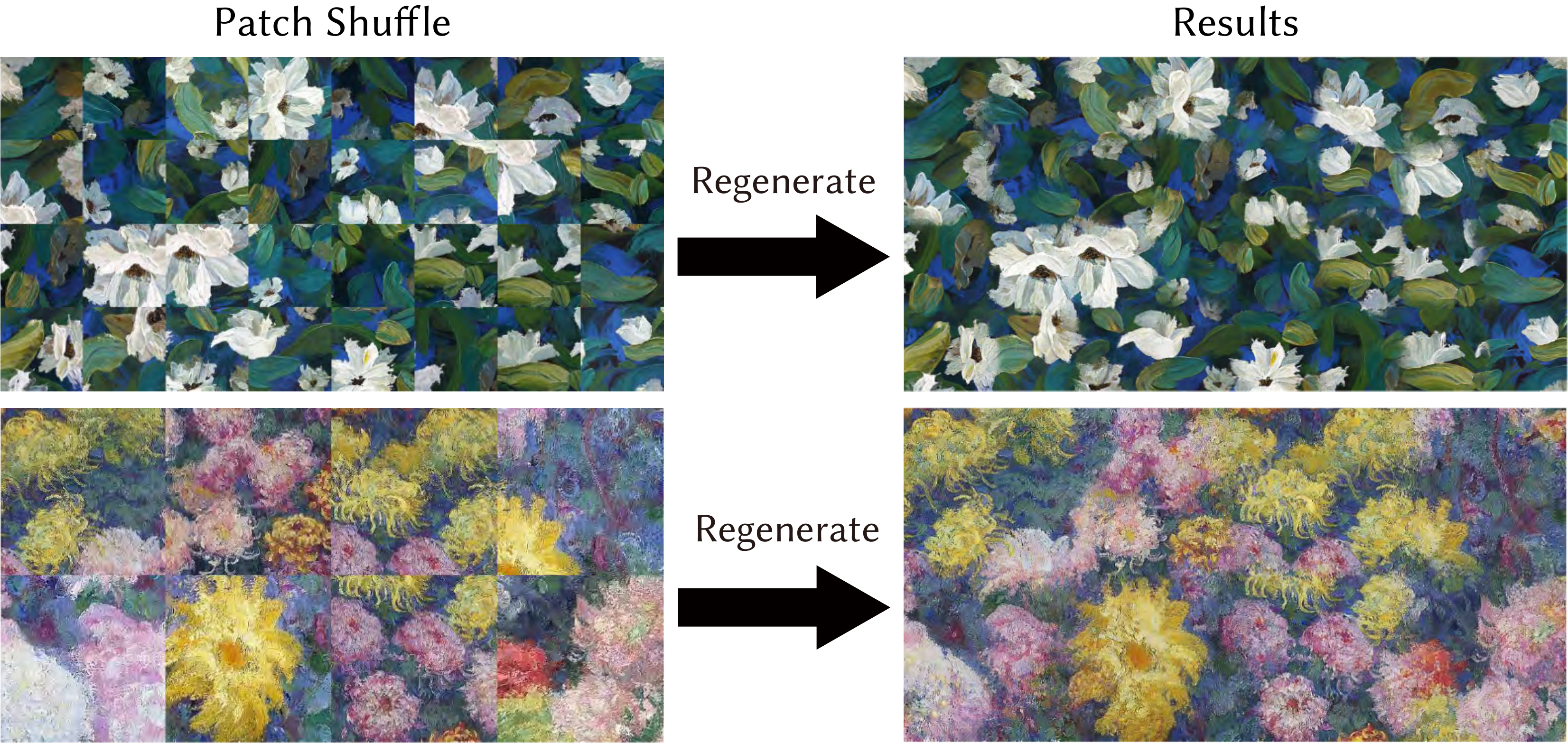}
    \caption{Patch-Shuffle: our method allows shuffling the original (sketch) patches to generate a novel texture. We show two cases with shuffled patches and the regenerated results.}
    \label{fig:shuffle}
\end{figure}

\subsection{Texture Shuffle}
For cases considered non-stationary textures, our framework allows shuffling the original sketch patches to regenerate novel textures.
The results of such an application are shown in Fig.~\ref{fig:shuffle}.

\begin{figure}[t!]
    \centering
    \includegraphics[width=\linewidth]{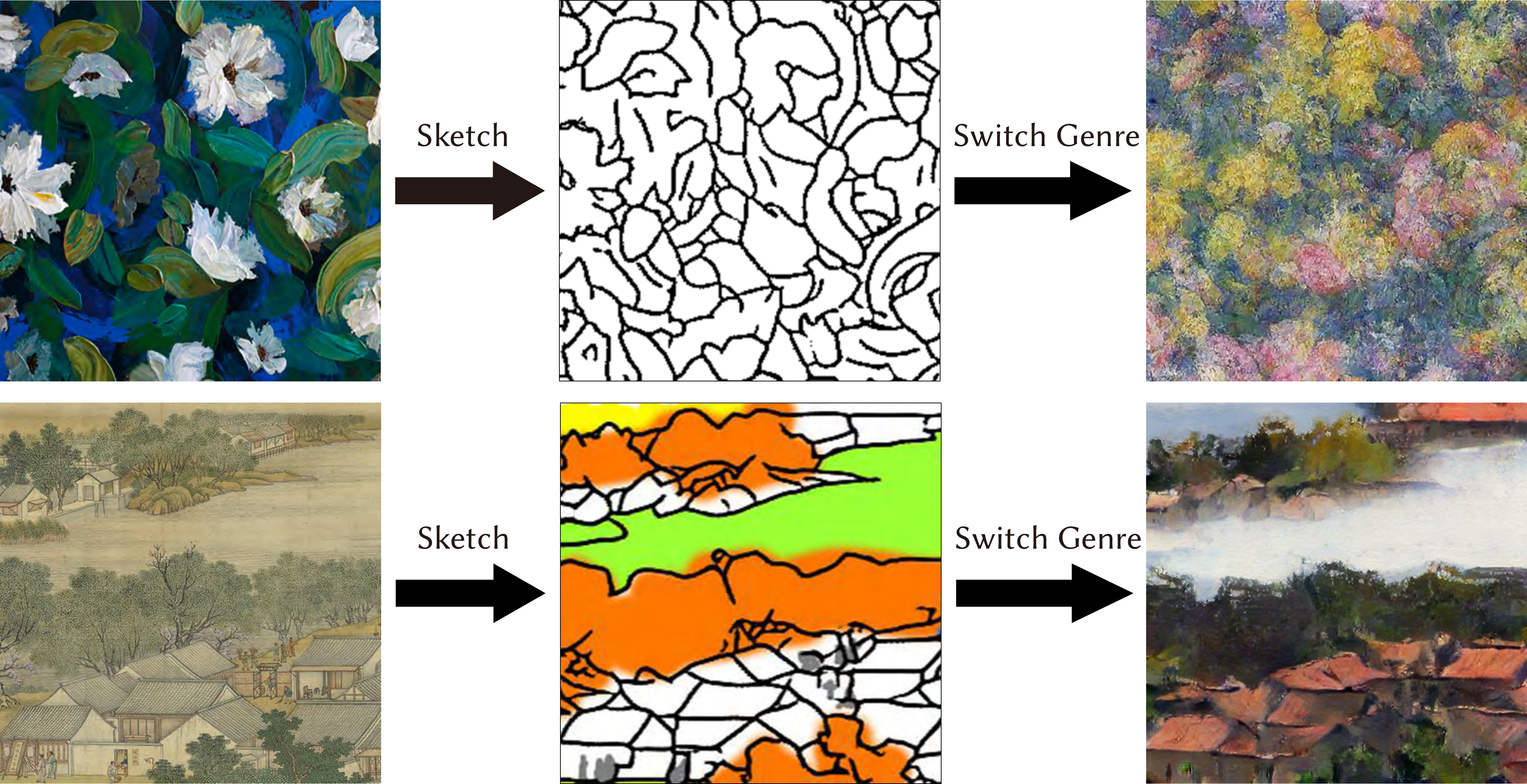}
    \caption{Genre switch. Given a painting, we generate the corresponding sketch according to the method described in Sec.~\ref{sec:SG}. Then, we feed the sketch image to another network of genres to achieve a digital painting in a switched genre.}
    \label{fig:switch}
\end{figure}

\begin{figure}[t]
\includegraphics[width=1.0\linewidth]{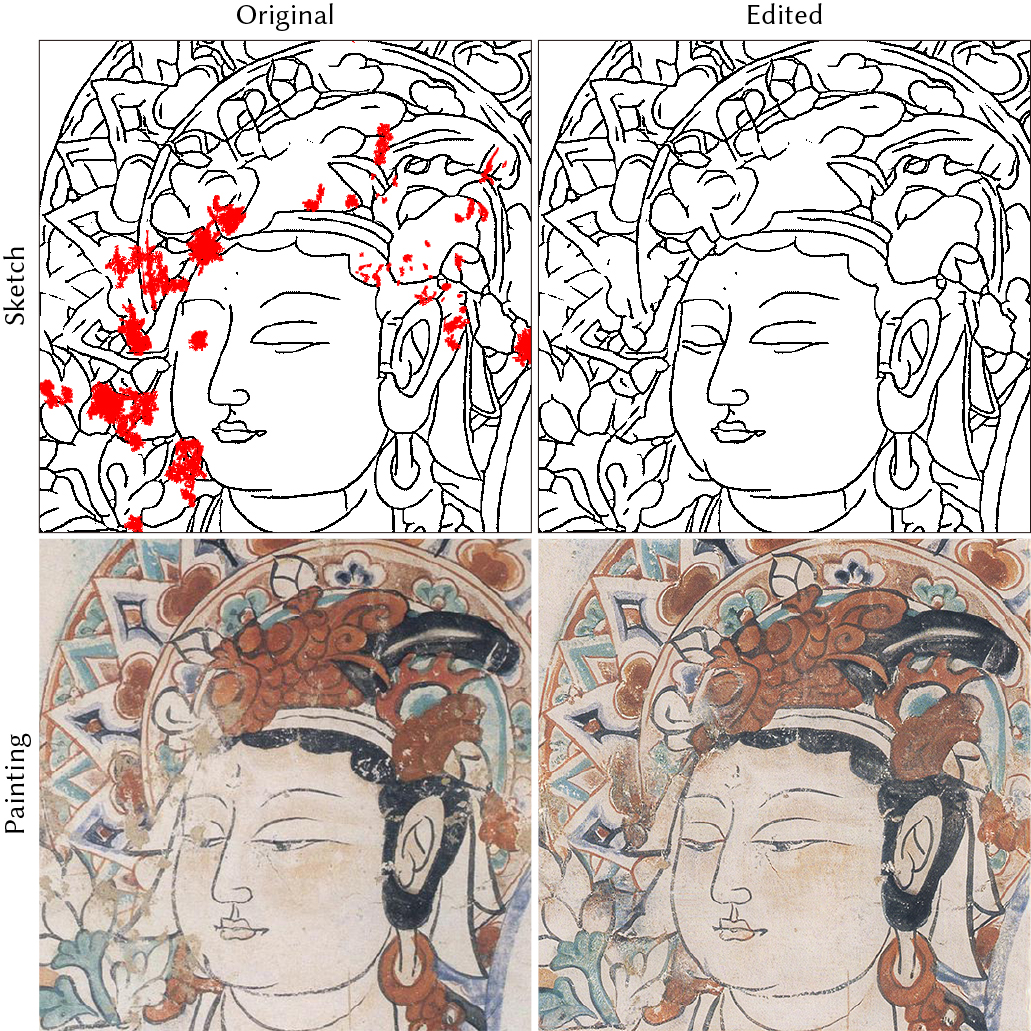}
\caption{Application: Mural restoration. The automatically generated sketch of the original painting is on the left, with weathered portions marked in red scribbles. With only a few strokes added to this sketch, a revitalized painting is produced (right), free of weathering effects.}
\label{fig:deweather}
\end{figure}

\subsection{Genre Switch}
For artistic paintings sharing similar compositions, we can also switch the genre from one to the other.
Fig.~\ref{fig:switch} demonstrates two examples of switching from one type of genre to the other.
Since our method constructs a one-way mapping from sketch to painting,
the proposed framework differs from style transfer.
With the intermediate domain of sketches, the recreation of artistic paintings is more controllable.

\begin{figure*}[t!]
    \centering
    \includegraphics[width=\linewidth]{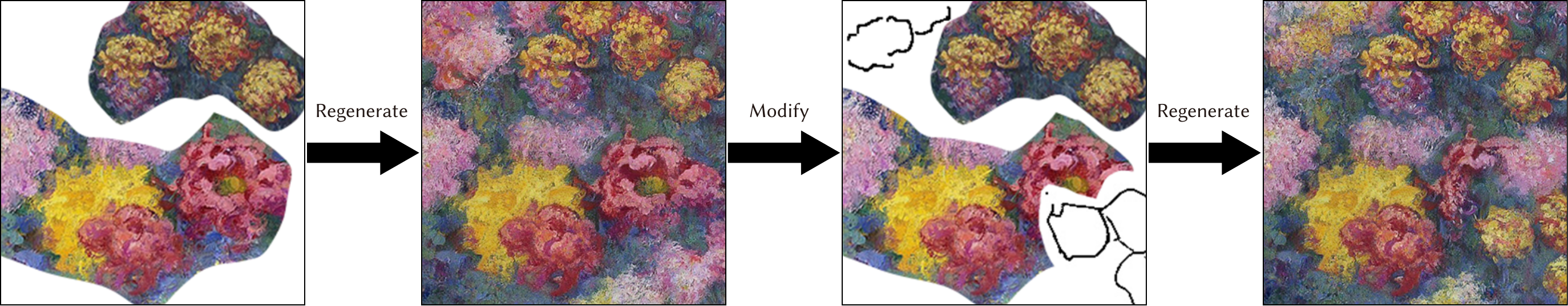}
    \caption{Cut-and-paste. Our model enables users to recompose novel digital paintings from their original parts. Even without a novel sketch, our framework produces novel textures in blank areas. Interactive sketches make generation more controllable.}
    \label{fig:app}
\end{figure*}

\begin{figure*}[t!]
\centering
\includegraphics[width=\linewidth]{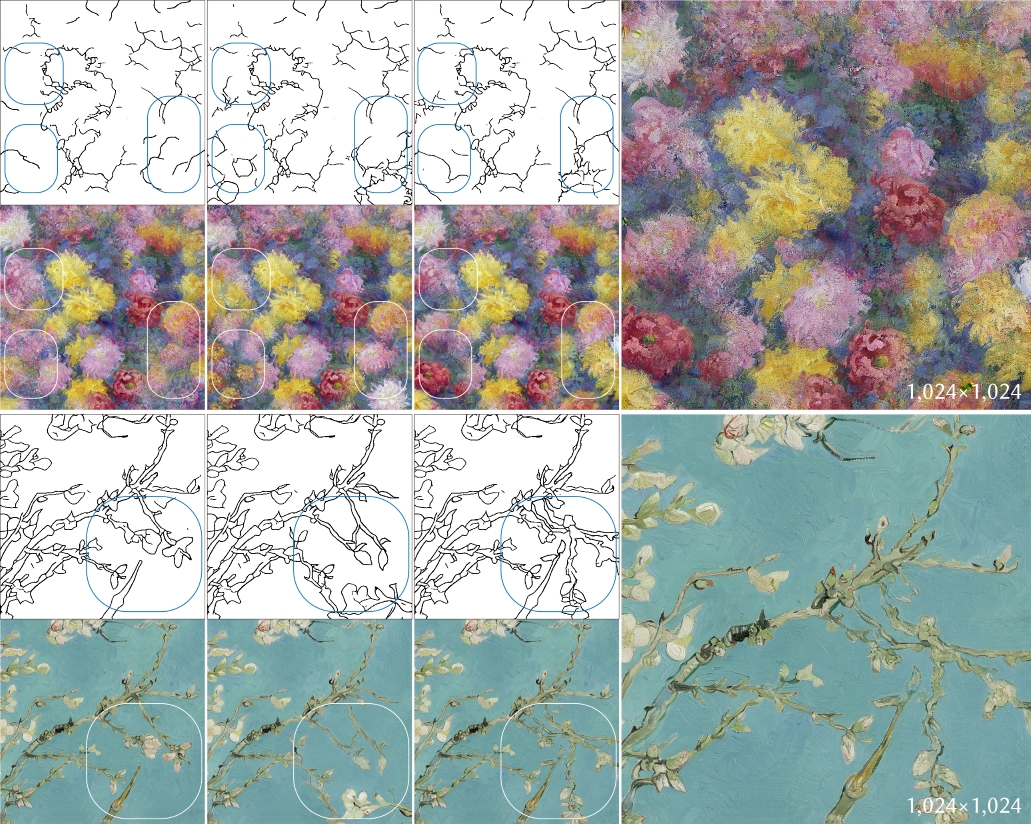}
\caption{Application: Locally controllable art generation. The boxed regions are edited interactively using the sketch input. The rightmost images provide close-ups of results in the third column.}
\label{fig:locally}
\end{figure*}

\begin{figure*}[tp!]
\centering
\includegraphics[width=\linewidth]{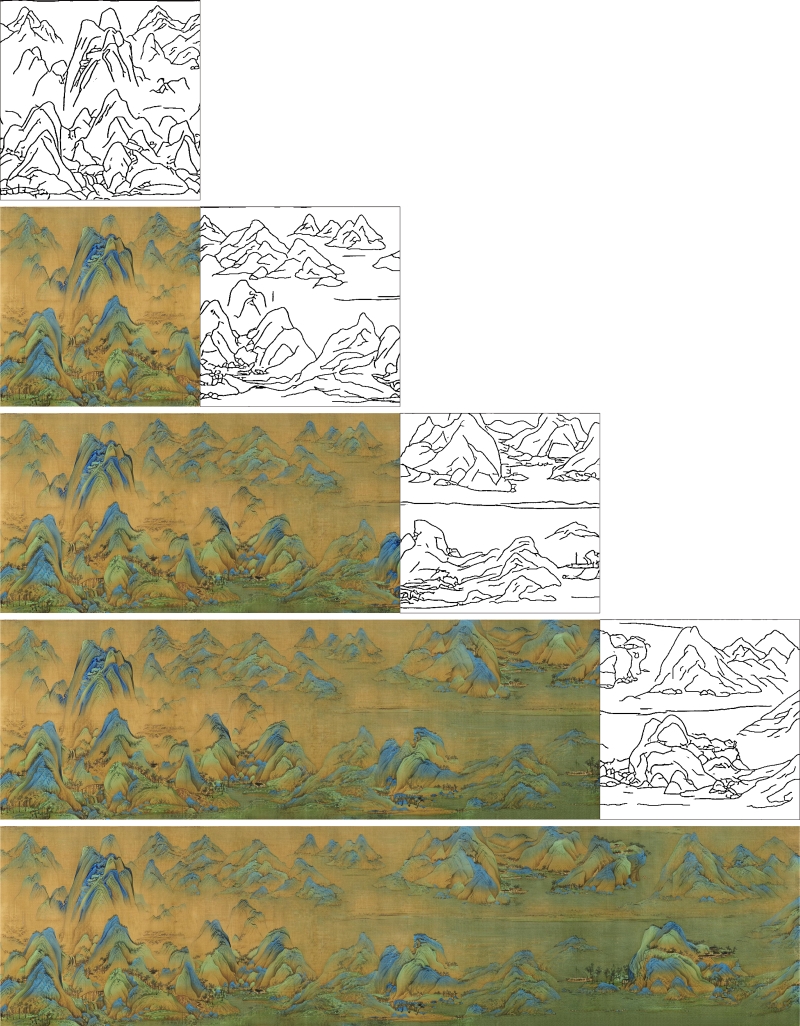}
\caption{Application: Progressive extension to be a very large artwork. These images show a sequence of progressive elaboration, where each sketch segment shares some overlapping elements with its neighboring segment from the synthesized result (from the second row to the bottom), bridging the sketches and the synthesized outcomes.}
\label{fig:blend}
\end{figure*}

\begin{figure*}[t!]
\includegraphics[width=1.0\linewidth]{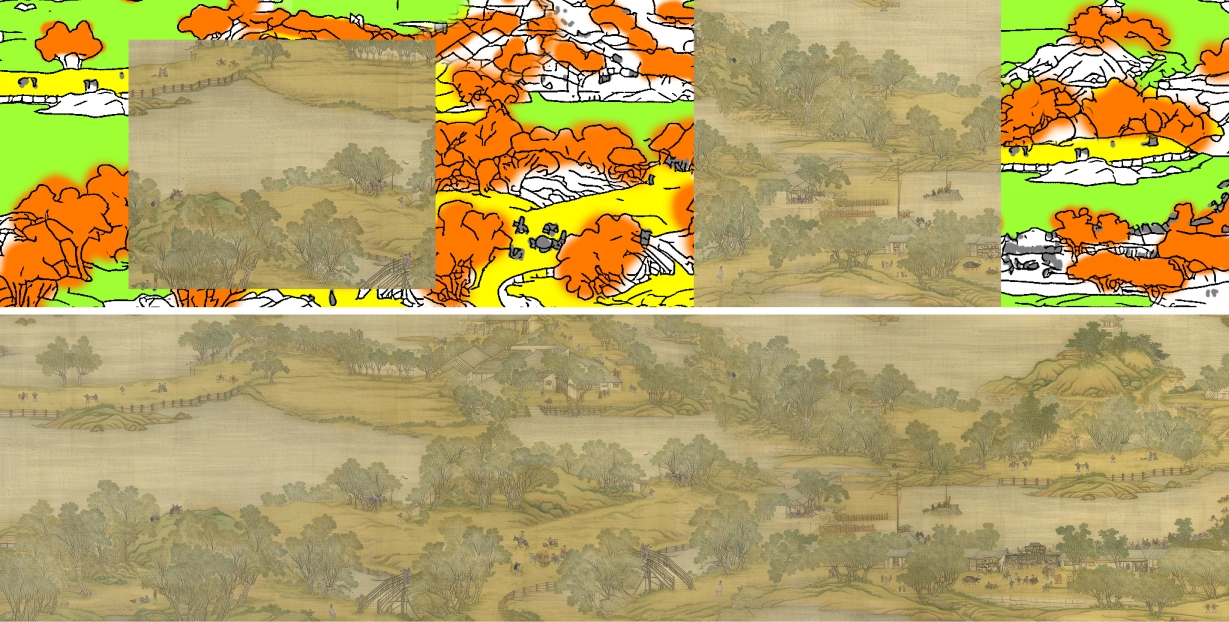}
\caption{Application: By placing the original painting patches onto an extended canvas filled with novel scribbled sketches input (upper), our method regenerates a new painting while preserving the original style and details (lower).
}
\label{fig:recompose}
\end{figure*}

\subsection{Other Applications}
In addition to the applications mentioned above, our method holds promise for various intriguing and significant applications. For example, as our proposed approach facilitates localized modifications to artworks, the model can be effectively employed in mural restoration, the model can be applied to mural restoration (see  Fig.~\ref{fig:deweather}) and other inpainting tasks and other inpainting tasks.

Fig.~\ref{fig:app} demonstrates a cut-and-paste example on the \emph{Chrysanthemums} data set. The generated result can be modified locally while maintaining a globally coherent appearance.

In Fig.~\ref{fig:locally}, we show locally controllable artwork generation. 
Fig.~\ref{fig:blend} shows a progressive process of large-scale artistic painting generation, as mentioned in our main paper. 
As demonstrated in Fig.~\ref{fig:recompose}, the model is capable of conditional generation with cropped painting patches and the users' free sketches.

\section{Conclusions and Limitations}
In this paper, we have presented a learning-based content-controllable painting recreation framework. Employing an end-to-end generative adversarial network and a multi-scale architecture enriched with an attention mechanism, our model demonstrates the capability to yield high-quality, high-definition outcomes.

Both quantitative and qualitative evaluations underscore the superiority of our proposed framework for generating artworks replete with intricate details compared over preceding methods that leverage style transfer or domain adaptation techniques to effortlessly produce images from sketches. Our framework not only empowers users to effortlessly craft or re-envision large-scale paintings using sketch-based inputs but also extends its applicability to other tasks like texture regeneration or high-quality genre transformation.

However, our approach also has certain limitations. For example, generating paintings enriched with nuanced semantic content necessitates sketching with semantic color masks. Since existing semantic segmentation methods may not be optimal for artistic paintings, the manual generation of these semantic masks remains a requisite, albeit requiring a relatively minor degree of manual effort. Furthermore, instances involving free inputs featuring fewer strokes and larger vacant spaces may inadvertently lead the model to generate suboptimal results.

\appendix


\subsection*{Declaration of competing interest}

The authors have no competing interests to declare that are relevant to the content of this article.\\

{\small
\bibliographystyle{CVMbib}
\bibliography{artgen_cvm}
}

\end{document}